\documentclass[11pt,a4paper]{article}
\usepackage[hyperref]{acl2021}
\usepackage{times}
\usepackage{latexsym}
\usepackage[printwatermark]{xwatermark}
\usepackage{amsmath,amssymb}
\usepackage{amsfonts}
\usepackage{booktabs}
\usepackage{multirow,multicol}
\usepackage{times}
\usepackage{pifont}
\usepackage{xcolor}
\usepackage{xspace}
\usepackage{csquotes}

\usepackage{textcomp}
\usepackage{fontawesome}
\usepackage{stfloats}
\usepackage{tabularx}
\usepackage{fontawesome}
\usepackage{algorithm}
\usepackage[ruled,vlined,boxed,algo2e]{algorithm2e}
\usepackage{scalerel}

\renewcommand{\Indentp}[1]{%
  \advance\leftskip by #1
  \advance\skiptext by -#1
  \advance\skiprule by #1}%
\renewcommand{\Indp}{\algocf@adjustskipindent\Indentp{\algoskipindent}}
\renewcommand{\Indm}{\algocf@adjustskipindent\Indentp{-\algoskipindent}}
\makeatother

\DeclareFontFamily{U}{skulls}{}
\DeclareFontShape{U}{skulls}{m}{n}{ <-> skull }{}
\newcommand{\argmax}{\mathop{\mathrm{argmax}}\limits}

\usepackage{microtype}

\aclfinalcopy 

\title{Meta-learning for downstream aware and agnostic pretraining}

\author{Hongyin Luo$^1$ $    $ Shuyan Dong$^2$$^*$ $    $ Yung-Sung Chuang$^3$  $    $ Shang-Wen Li$^2$\thanks{Work in progress. * Work is not related to employment at Amazon.} \\
  $^{1}$MIT CSAIL, $^{2}$Amazon AI, $^{3}$National Taiwan University \\
  {\tt \{hyluo\}@mit.edu, \{shuyand,shangwel\}@amazon.com,\{chuangyungsung\}@gmail.com}\\
  }

\date{}

\begin{document}
\maketitle
\begin{abstract}
Neural network pretraining is gaining attention due to its outstanding performance in natural language processing applications. However, pretraining usually leverages predefined task sequences to learn general linguistic clues. The lack of mechanisms in choosing proper tasks during pretraining makes the learning and knowledge encoding inefficient. We thus propose using meta-learning to select tasks that provide the most informative learning signals in each episode of pretraining. With the proposed method, we aim to achieve better efficiency in computation and memory usage for the pretraining process and resulting networks while maintaining the performance. In this preliminary work, we discuss the algorithm of the method and its two variants, downstream-aware and downstream-agnostic pretraining. Our experiment plan is also summarized, while empirical results will be shared in our future works.
\end{abstract}

\section{Introduction and related works}

Recently neural network pretraining, such as BERT \citep{devlin2018bert}, OpenAI GPT \citep{radford2018improving}, XLNet \citep{yang2019xlnet}, and ELECTRA \citep{clark2019electra} has yielded significant performance improvement for various downstream applications in natural language processing. Usually, semantic and syntactic information is implicitly learned from the co-occurrence of words and sentences in datasets and encoded in the networks for general purposes. The entire networks or part of their layers are then finetuned for the downstream usage. To better leverage training corpora and discover all valuable information, ERNIE 2.0 \citep{sun2020ernie} proposes a sequential multitask pretraining framework containing a variety of word-, structure-, and semantic-aware pretraining tasks. Although increasing the diversity of pretraining tasks continually improves model performance accuracy, the lack of mechanisms in selecting and scheduling tasks makes the training process, model serving, and memory footprint bulky. The model expressivity has to grow accordingly to encode information from all the tasks generally.

To address the challenge, we propose a framework using meta-learning to schedule tasks and make pretraining more efficient in this preliminary paper. Meta-learning is commonly used in learning model architectures, hyperparameters, or learning algorithms that are generalizable across tasks through episodic training \citep{finn2017model, vinyals2016matching, snell2017prototypical, luo2020prototypical, li2021meta}. Hence, we leverage meta-learning to formulate each pretraining batch as an episode, and select the pretraining task offering the most helpful training signals in the episode, instead of following predefined or sequential order of tasks, to update the model parameters. With the formulation, we explore two variants, downstream-aware and downstream-agnostic pretraining, that utilize loss evaluating training signals with or without labels in downstream applications. The episodic task selection is expected to encode lexical, syntactic, and semantic information across various tasks more efficiently. In this preliminary work, we introduce the detailed algorithm, implementation challenges and corresponding solutions, and experiment settings. We will share empirical results later.

\section{Method and experiment plan}

To make multitask pretraining more efficient, we propose a framework of utilizing meta-learning to schedule tasks used for network pretraining. We summarize the framework step-by-step in Algorithm \ref{alg:meta-pretrain}. Like many previous meta-learning works, we use two sets of tasks, source $S$ and target $T$, to build episodes for pretraining. $S$ are different pretraining tasks such as masked language modeling, sentence reordering, and dialog response selection; $T$ can be similar tasks to $S$ or a set of downstream applications. Each \texttt{while} loop in the algorithm defines an episode and is analogous to parameter update with $N$ batches in ordinary pretraining. However, before iterating through each batch of examples from the pretraining task $\hat{s}$ and updating model parameter $\theta$ (line 11 and 12), we modify an across-task training technique (line 3 to 9) commonly used in meta-learning \citep{finn2017model, snell2017prototypical} to sample subtasks (i.e., a batch of examples) from $S$ and $T$ for selecting the task $\hat{s}$ to pretrain on. We first obtain $\tau^T$ by sampling $M$ subtasks for each task in $T$ in line 3. Then, we iterate through each task $s$ in $S$, sample $N$ subtasks, $\tau_i^s$, for $s$, and evaluate the utility of $s$, $u(s)$, with $\tau_i^s$ and $\tau^T$. The utility is computed by updating $\theta$ with gradient evaluated on $\tau_i^s$ and scoring the updated parameter $\theta'$ with $\tau^T$. By selecting the source task yielding the largest utility to pretrain on\footnote{Similar to approaches in reinforcement learning, the epsilon-greedy algorithm can be used to choose between exploration and exploitation randomly.}, we expect more efficient network pretraining.

We employ the equation, $scorer(\theta', \tau^T) = -\sum_{j, t} L_{\tau_j^t}(f(\theta'))$, to measure the utility of tasks. As indicated in the equation, we prefer the source task resulting in updated parameters ($\theta'$) that yield smaller summed losses over target subtasks. With this formulation, it is crucial to choose appropriate target tasks to compute the utility and thus guide the pretraining. Ideally, the target tasks should be as relevant to downstream applications as possible or even provide a similar labeling schema, and we refer to such setting as downstream-aware pretraining. Sometimes a generalizable pretrained network is preferred. In this case, we have $T = S$ for downstream-agnostic pretraining. Both settings will be explored in our experiments.


For our experiment, we will adopt common Transformer architectures such as BERT \citep{devlin2018bert} and RoBERTa \citep{liu2019roberta} as the backbone models and examine the performance and pretraining efficiency when the proposed method is applied. We plan to explore source pretraining tasks based on ERNIE 2.0 \citep{sun2020ernie} as the work provides good coverage of tasks for word-, structure-, and semantic-aware pretraining signals. Masked passage retrieval \citep{glass2020span}, text generation, and unsupervised question answering \citep{luo2021cooperative} tasks will also be investigated to provide generation and passage level pretraining losses. 

\begin{algorithm}[H]
 \caption{Meta-learning based network pretraining}
 \label{alg:meta-pretrain}
 \textbf{Inputs:} Set of all source and target tasks $S, T$; data distribution over each source and target task $p^s(\tau), p^t(\tau)$; $L_{\tau}$, the loss function evaluated on a subtask $\tau$ sampled from $p^s(\tau)$ or $p^t(\tau)$; learning rates $\alpha$, $\lambda$, number of batches $M, N$. \\
 \textbf{Output:} Optimized network parameters $\theta$. \\
 \nl Initialize $\theta$; \\
 \nl \While {not done} {
   \nl Sample $M*|T|$ target subtasks (i.e., batches) $\tau^T = \bigcup\limits_{t=T}\tau_j^t$, where $\tau_j^t \sim p^t(\tau)$ and $j = 1...M$; \\
   \nl \For{$s \in S$}{
    \nl Sample $M$ subtasks $\tau_i^s \sim p^s(\tau), i = 1...M$; \\
    \nl $u(s) = 0$; \\
    \nl \For{all $\tau_i^s$}{
     \nl Compute gradient for fast adaptation: $\theta' = \theta - \alpha \nabla_{\theta} L_{\tau_i^s}(f(\theta))$; \\
     \nl Evaluate updated parameters on target subtasks: $u(s)$ += $scorer$($\theta'$, $\tau^T$)
    }
   }
   \nl $\hat{s} = \argmax_s u(s)$ \\
   \nl Sample $N$ subtasks $\tau_i^{\hat{s}} \sim p^{\hat{s}}(\tau), i = 1...N$; \\
   \nl \For{all $\tau_i^{\hat{s}}$}{
    Compute gradient and update parameters: $\theta = \theta - \lambda \nabla_{\theta} L_{\tau_i^{\hat{s}}}(f(\theta))$; \\
   }
 }
\end{algorithm}

We will focus on four applications for the downstream tasks and the target pretraining tasks in the downstream-aware setting: relation extraction, semantic role labeling, general language and sentence understanding, and machine reading comprehension. In this setting, additional prediction heads will be added to the backbone network separately for different tasks. These heads are pretrained (in the downstream-aware setting) and finetuned (in both downstream-aware and downstream-agnostic settings) for evaluation. We will employ CoNLL04 \citep{roth2004linear}, CoNLL-2005 \citep{carreras2005introduction}, GLUE \citep{wang2018glue}, and SQuAD \citep{rajpurkar2016squad}, respectively, to obtain learning signals for the target pretraining tasks of relation extraction, semantic role labeling, general language and sentence understanding, and machine reading comprehension. Correspondingly, ACE2005 \citep{walker2006ace}, CoNLL 2012 \citep{pradhan2012conll}, SuperGLUE \citep{wang2019superglue} (excluding those tasks already seen in GLUE), and Natural Questions \citep{kwiatkowski2019natural} will be used as the downstream applications to evaluate pretrained networks in each of the four tasks.



\section{Conclusion}
In this work, we propose a meta-learning-based downstream-aware and downstream-agnostic pretraining method, where pretraining tasks are selected episodically from a list of candidate tasks to improve pretraining efficiency while maintaining the performance. We summarize the algorithm and our experiment plan. Empirical results will be shared in our future works.

\bibliography{acl2021}
\bibliographystyle{acl_natbib}




\end{document}